\documentclass{bytedance_seed}

\usepackage{amssymb}
\usepackage{makecell}
\usepackage{multirow}
\usepackage{newtxtext,newtxmath}
\usepackage{float}

\title{Partial FC: Training 10 Million Identities on a Single Machine}
\author{Xiang An, Xuhan Zhu, Yang Xiao, Lan Wu, Ming Zhang,\\
  Yuan Gao, Bin Qin, Debing Zhang, Ying Fu, Jiankang Deng\\[12pt]
  \href{https://www.anxiangsir.com/pages/partial_fc}{Project Page}}

\abstract{
Training face recognition models with millions of identities is challenging
because classifier storage, logit memory, and computation grow linearly with the
number of classes, eventually making full softmax impractical even when the
backbone itself fits comfortably in memory. We present Partial FC (PFC), a
scalable approximation to large-class softmax that preserves every positive
class center while activating only a sampled subset of negative centers in each
mini-batch. This asymmetric treatment retains every target term while avoiding
exhaustive interaction with millions of mostly uninformative negatives. Our
distributed implementation partitions the classifier across GPUs and samples
within each owned shard, so sampling reduces local matrix multiplication and
logit storage while sharding eliminates class-gradient synchronization across
workers. Together, these properties reduce GPU-resident classifier memory,
computation, and class-dependent communication without feature-based
hard-negative retrieval. End-to-end system benchmarks demonstrate efficient
scaling to massive class spaces, including 64 million classes on a single
eight-GPU machine. Across large-scale face-recognition datasets, moderate
sampling rates maintain competitive recognition accuracy while substantially
improving training efficiency. Our best PFC configurations achieve 97.2\% TAR on
IJB-C at FAR $=10^{-4}$ and 94.0\% TAR on ICCV21-MFR at FAR $=10^{-6}$. Beyond
clean training data, PFC is robust to inter-class conflicts, label noise, and
long-tailed identity distributions: under 40\% label noise, PFC with conflict
filtering raises ICCV21-MFR TAR from 43.9\% to 80.2\%, while on long-tailed data
PFC improves TAR from 87.4\% to 92.0\%. These results establish
positive-preserving negative sampling as an effective foundation for scalable,
accurate, and robust identity classification. }
\begin{document}
\maketitle

\section{Introduction}

Large-scale face recognition learns discriminative embeddings by classifying
each image among millions of identities. Training collections have progressed
from Celeb-500K to WebFace260M, whose largest setting contains millions of
identities and hundreds of millions of images
\cite{cao2018celeb,zhu2021webface260m}. In parallel, angular and cosine margin
losses---SphereFace, AM-Softmax, CosFace, and ArcFace---have made normalized
softmax the dominant training paradigm
\cite{liu2017sphereface,wang2018additive,wang2018cosface,deng2019arcface}.
Subsequent objectives such as GroupFace, Circle Loss, and CurricularFace further
improve feature discrimination or optimization
\cite{kim2020groupface,sun2020circle,huang2020curricularface}, but retain the
same full-class comparison. Their accuracy therefore comes with a systems cost:
for embedding dimension $d$, class count $C$, and batch size $B$, full softmax
maintains a $d\times C$ classifier and materializes $B\times C$ logits. Both
storage and computation grow linearly with identity scale, and the
classification head can become the bottleneck even when the backbone remains
tractable.

Existing acceleration strategies address only part of this problem. HF-Softmax
performs dynamic class selection with a random hash forest, reducing arithmetic
by retrieving nearby active centers from the feature space, but introducing
additional RAM storage and search cost \cite{zhang2018accelerated}. Softmax
Dissection separates intra-class and inter-class objectives to avoid redundant
negative computation, but is specialized to its decomposed loss
\cite{he2020softmax}. ArcFace-style model parallelism shards the classifier
across devices and avoids synchronizing the full weight gradient
\cite{deng2019arcface}; however, it still evaluates every class, so local logits
grow with the global batch and eventually dominate memory. Conventional data
parallelism has the complementary cost of replicating the classifier and
synchronizing its gradients \cite{li2014scaling}. A scalable solution must
reduce the active class dimension itself without losing the target supervision
that makes classification effective.
\begin{figure}[tbp]
  \centering
  \includegraphics[width=0.94\textwidth]{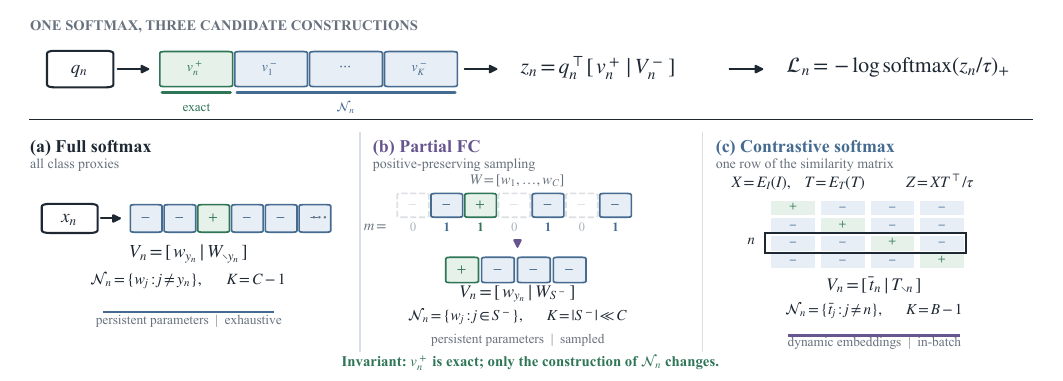}
  \caption{Positive-preserving candidate softmax. Full softmax scores all
    persistent class centers, whereas Partial FC keeps the positive center and
    samples only negative centers. Contrastive learning has the same row-wise
    softmax structure, but its paired positive and in-batch negatives are
    dynamically produced by the opposite encoder.}
\label{fig:candidate_matching}
\end{figure}

Our key observation is that positive and negative classes play fundamentally
different roles. The positive center provides the indispensable target term that
attracts an embedding toward its identity, whereas each negative contributes
only through its softmax probability. In classifiers with millions of
identities, most negatives are already well separated and have negligible
gradients. This asymmetry suggests a simple principle: preserve every positive
exactly and approximate only the negative denominator.

Partial FC (PFC) implements this principle by separating permanent class
ownership from transient activation \cite{an2022killing}. The classifier is
partitioned across devices; each rank retains every positive center owned by its
shard and fills a fixed activation budget with randomly sampled negatives.
Sampling reduces local matrix multiplication and logit storage, while permanent
ownership keeps classifier updates local and removes class-dependent gradient
synchronization. Unlike feature-based class selection, PFC requires no
nearest-center search. The same partial activation also lowers the frequency
with which mislabeled duplicate identities appear as negatives, and an optional
conflict filter suppresses abnormally similar negative logits under severe label
noise.

The resulting system scales classifier training to 64 million classes on a
single eight-GPU machine. On real face-recognition benchmarks, our best PFC
configurations achieve 97.2\% TAR on IJB-C at FAR $=10^{-4}$ and 94.0\% TAR on
ICCV21-MFR at FAR $=10^{-6}$. PFC also improves robustness when web-scale
supervision is imperfect: with 40\% label noise, conflict filtering raises
ICCV21-MFR TAR from 43.9\% to 80.2\%, and on long-tailed training data PFC
improves TAR from 87.4\% to 92.0\%.

Figure~\ref{fig:candidate_matching} places this mechanism in a broader
candidate-matching view. A proxy classifier scores an image against persistent
class centers, while a contrastive objective such as CLIP scores it against
dynamically encoded candidates \cite{radford2021learning}. Both contain one
indispensable positive and a potentially large negative denominator, and
negative-prototype sampling has also proved effective for million-pseudo-class
representation learning \cite{an2023unicom}. The shared principle is
positive-preserving approximation, but the systems differ: PFC directly updates
persistent centers, whereas contrastive candidates are transient and route
gradients through both encoders. This work therefore develops and evaluates
distributed sampling for persistent classifier centers; extending the
implementation to dynamic contrastive candidates remains future work.

Our contributions are threefold:
\begin{itemize}
  \item[1)] We present a candidate-softmax view that identifies
    positive-preserving negative sampling as a shared approximation principle for
    classification and contrastive objectives, while making their optimization
    and systems boundary explicit.
  \item[2)] We develop a distributed PFC implementation that combines permanent
    class ownership with batch-level negative sampling, reducing active GPU
    classifier storage, logits, computation, and class-dependent communication
    without feature retrieval.
  \item[3)] We validate PFC through system scaling to 64 million classes,
    competitive large-scale face-verification accuracy, and substantial
    robustness gains under inter-class conflict, label noise, and long-tailed
    training data.
\end{itemize}

\section{Related Work}

\subsection{Large-Scale Face Recognition}
Deep face recognition represents an image as a compact embedding and compares it
by distance or similarity. FaceNet established the triplet-loss embedding
paradigm for recognition, verification, and
clustering~\cite{schroff2015facenet}, while classification-based systems trained
on CASIA-WebFace and VGGFace demonstrated the strength of identity supervision
at larger data scales~\cite{yi2014learning,parkhi2015deep}. The available
supervision has since expanded substantially: Celeb-500K contains roughly half a
million identities~\cite{cao2018celeb}, and WebFace260M provides benchmark
subsets with millions of identities and hundreds of millions of
images~\cite{zhu2021webface260m}. MegaFace complements these training resources
with a million-distractor evaluation protocol for recognition at
scale~\cite{kemelmacher-shlizerman2016the}.

Normalized margin softmax has become the dominant objective for learning
discriminative face embeddings. SphereFace introduces a multiplicative angular
margin~\cite{liu2017sphereface}; AM-Softmax and CosFace use additive margins in
cosine space~\cite{wang2018additive,wang2018cosface}; and ArcFace applies an
additive angular margin with a direct geometric
interpretation~\cite{deng2019arcface}. Later methods improve the representation
or optimization from complementary directions. GroupFace learns latent
group-aware representations~\cite{kim2020groupface}, Circle Loss adaptively
weights pair similarities~\cite{sun2020circle}, and CurricularFace adjusts the
emphasis on hard examples over training~\cite{huang2020curricularface}. These
methods substantially improve discrimination, but their class-level
implementations still compare embeddings against a classifier whose size grows
with the number of identities. The present work is orthogonal to the choice of
margin loss: it targets the class dimension of this computation.

\subsection{Efficient Large-Class Softmax}
For a $C$-class classifier with embedding dimension $d$ and batch size $B$, full
softmax stores $O(dC)$ classifier parameters and evaluates $O(BC)$ logits.
Several methods reduce this cost by selecting or restructuring classes.
HF-Softmax dynamically retrieves active classes with a randomized hash forest in
the embedding space~\cite{zhang2018accelerated}. This focuses computation on
nearby classes, but requires maintaining and searching an index over class
centers. Softmax Dissection separates intra-class and inter-class components and
reduces redundant inter-class computation within its decomposed embedding
objective~\cite{he2020softmax}. Class-based approximations have also been
studied for fast maximum-entropy training~\cite{goodman2001classes}. These
approaches show that exhaustive normalization is not always necessary, but they
either introduce a retrieval/decomposition mechanism or do not enforce the
positive-preserving sampling rule required by margin-based face classifiers.

Partial FC instead retains every positive center and randomly activates only a
subset of negatives~\cite{an2022killing}. This distinction is important because
omitting a target class removes its attractive supervision, whereas omitting a
well-separated negative often removes only a small probability term. PFC does
not require nearest-center search and remains compatible with normalized softmax
variants such as SphereFace, CosFace, and ArcFace. Relative to its earlier
formulation, this work makes the batch-shared sampling distribution explicit,
analyzes its relation to full softmax, and develops the distributed systems view
in terms of permanent class ownership and transient activation.

\subsection{Distributed Classifier Training}
Distributed data parallelism replicates model state and synchronizes gradients,
following the broader parameter-server and distributed-SGD
paradigm~\cite{li2014scaling}. This design is effective when the model is
moderate, but replicating a million-class classifier on every worker is
expensive. Class-sharded model parallelism addresses classifier storage by
assigning disjoint columns to different ranks. ArcFace demonstrates this
strategy for million-identity training: each rank computes logits for its local
class shard, while collective reductions recover the exact global
softmax~\cite{deng2019arcface}. Classifier gradients remain local, avoiding an
all-reduce over the full weight matrix.

Sharding alone, however, does not reduce the number of evaluated classes. Every
rank processes the gathered global batch against all centers in its shard, so
local logit storage grows with both global batch size and class count. Adding
workers increases classifier capacity, but can simultaneously increase the
global-batch logits handled by each rank. PFC combines the two complementary
mechanisms: permanent ownership distributes the classifier, while
positive-preserving activation reduces each local shard before matrix
multiplication. As a result, classifier-gradient communication remains local and
the communicated feature payload does not grow with $C$.

\section{Method}

In this section, we first derive the candidate-matching structure shared by
softmax classification and contrastive learning, while distinguishing how their
candidates are represented and updated. We then analyze exact class-sharded
model parallelism and its memory limits before introducing the
positive-preserving approximation and its distributed implementation.

\subsection{Softmax Classification and Contrastive Learning as Candidate Matching}
\paragraph{Why softmax weights become class centers.} Consider a linear
$C$-class softmax classifier with image embedding $x_i$, label $y_i$, and
classifier $W=[w_1,\ldots,w_C]$. Omitting the bias for clarity, its logits,
probabilities, and per-example loss are
\begin{equation}
f_{i,j}=w_j^Tx_i,\qquad
p_{i,j}=\frac{\exp(f_{i,j})}{\sum_{c=1}^{C}\exp(f_{i,c})},\qquad
\ell_i=-\log p_{i,y_i}.
\label{eq:proxy_softmax}
\end{equation}
The score gradient, weight gradient, and an SGD step with learning rate $\eta$
are
\begin{equation}
\frac{\partial\ell_i}{\partial f_{i,j}}
=p_{i,j}-\mathbf{1}[j=y_i],\qquad
\frac{\partial\ell_i}{\partial w_j}
=\bigl(p_{i,j}-\mathbf{1}[j=y_i]\bigr)x_i,
\qquad
\Delta w_j
=\eta\bigl(\mathbf{1}[j=y_i]-p_{i,j}\bigr)x_i.
\label{eq:proxy_update}
\end{equation}
For the target class, $\Delta w_{y_i}=\eta(1-p_{i,y_i})x_i$ attracts the weight
toward a feature of that class. For every non-target class, $\Delta w_j=-\eta
p_{i,j}x_i$ repels its weight from the feature, with stronger competitors
receiving larger updates. Accumulated over many mini-batches, each persistent
column $w_j$ is therefore attracted by examples of class $j$ and repelled by
examples of other classes. This makes it a discriminatively learned \emph{class
proxy}, commonly called a class center, rather than the arithmetic mean of the
class features.

Modern face classifiers normalize both vectors and use $f_{i,j}=s\bar w_j^T\bar
x_i=s\cos\theta_{i,j}$, optionally modifying the target logit with a
margin~\cite{wang2018cosface,liu2017sphereface,deng2019arcface}. In this
setting, $\bar w_j$ represents a class direction on the unit hypersphere.
Differentiating through normalization projects the update in
Eq.~\ref{eq:proxy_update} onto the tangent space of that sphere, and a margin
changes the target coefficient, but the same positive-attraction and
negative-repulsion interpretation remains.

\paragraph{PFC updates persistent class centers.} PFC does not introduce a
different kind of classifier weight. It retains the same persistent class
centers but activates only a subset in each iteration. Let $\mathcal{S}$ be the
active class set shared by a mini-batch, containing the target $y_n$ of every
example $n$, and define
\begin{equation}
\widehat p_{n,j}
=\frac{\exp(f_{n,j})}{\sum_{c\in\mathcal{S}}\exp(f_{n,c})},
\qquad j\in\mathcal{S}.
\label{eq:pfc_probability_overview}
\end{equation}
For plain softmax, the per-example contributions to the classifier update are
\begin{equation}
\Delta w_j=
\begin{cases}
\eta(1-\widehat p_{n,y_n})x_n,
& j=y_n,\\
-\eta\widehat p_{n,j}x_n,
& j\in\mathcal{S}\setminus\{y_n\},\\
0,
& j\notin\mathcal{S}.
\end{cases}
\label{eq:pfc_center_update}
\end{equation}
The mini-batch contributions are summed, so a center may be positive in one row
and negative in others; normalized margin softmax additionally applies its
scale, margin derivative, and tangent-space projection.
Equation~\ref{eq:pfc_center_update} shows the key asymmetry: all positives
receive attractive updates, but only sampled negatives receive repulsive
updates. Unsampled centers remain in the classifier and optimizer state for
later iterations. Since $\widehat p_{n,j}$ is normalized over $\mathcal{S}$,
these are not generally unbiased full-softmax gradients; random resampling
instead provides continually refreshed negative coverage.

\paragraph{CLIP updates encoders through dynamic candidates.}
CLIP~\cite{radford2021learning} has the same row-wise score structure but no
persistent candidate matrix. Given $B$ image--text pairs, its encoders produce
normalized embeddings $\bar x_i$ and $\bar t_i$. For the image-to-text
direction, define
\begin{equation}
q_{i,j}^{I\rightarrow T}
=\frac{\exp(\bar x_i^T\bar t_j/\tau)}
{\sum_{k=1}^{B}\exp(\bar x_i^T\bar t_k/\tau)},
\qquad
\ell_i^{I\rightarrow T}=-\log q_{i,i}^{I\rightarrow T}.
\label{eq:clip_i2t}
\end{equation}
The paired text is positive and the others are in-batch negatives.
Differentiating with respect to the normalized text embeddings gives
\begin{equation}
\frac{\partial\ell_i^{I\rightarrow T}}{\partial\bar t_j}
=\frac{1}{\tau}
\bigl(q_{i,j}^{I\rightarrow T}-\mathbf{1}[j=i]\bigr)\bar x_i.
\label{eq:clip_candidate_gradient}
\end{equation}
Unlike a class center, $\bar t_j$ is transient and passes its gradient to the
shared text encoder by the chain rule,
\begin{equation}
\nabla_{\theta_T}\mathcal L
=\sum_{j=1}^{B}J_{\theta_T}(\bar t_j)^T
\nabla_{\bar t_j}\mathcal L.
\label{eq:clip_encoder_update}
\end{equation}
CLIP regenerates candidate embeddings each iteration and also optimizes the
symmetric text-to-image direction, $\mathcal L_{\mathrm{CLIP}}=(\mathcal
L_{I\rightarrow T}+\mathcal L_{T\rightarrow I})/2$, so both encoders receive
gradients.

\paragraph{Shared rule and important boundary.} At the score level,
Eqs.~\ref{eq:proxy_update} and~\ref{eq:clip_candidate_gradient} share the same
form: the positive receives the target term, while each negative contributes its
softmax probability. Both instantiate
\begin{equation}
\mathcal L_i(u_i,v_i^+)
=-\log
\frac{\exp\bigl(\operatorname{sim}(u_i,v_i^+)/\tau\bigr)}
{\exp\bigl(\operatorname{sim}(u_i,v_i^+)/\tau\bigr)
+\sum_{v\in\mathcal N_i}\exp\bigl(\operatorname{sim}(u_i,v)/\tau\bigr)}.
\label{eq:unified_candidate_loss}
\end{equation}
This common form motivates retaining the positive while sampling only the
negative denominator. The update carrier remains different: PFC directly updates
persistent class centers, whereas CLIP updates shared encoders through transient
sample features. The objectives therefore share an approximation principle, not
a storage or optimization mechanism. Extending the implementation below beyond
persistent class centers requires communication of dynamic candidates and
gradient routing through both encoders.

Having established positive-preserving negative sampling as the approximation
principle, we now focus on classifiers with persistent class centers, where the
candidate bank can be explicitly sharded and sampled. We first quantify the
limitations of full-class model parallelism and then derive the distributed
Partial FC implementation.

\subsection{Problem Formulation}
\subsubsection{Model Parallelism}

For a batch $X\in\mathbb{R}^{B\times d}$, full softmax computes $Z=XW$ and
$P=\operatorname{softmax}(Z)$, with $G=\frac{1}{B}(P-Y)$, $\nabla W=X^TG$, and
$\nabla X=GW^T$. The $B\times C$ logits and $d\times C$ classifier are
impractical when $C$ is in the millions. We therefore shard $W=[W_1,\ldots,W_k]$
by classes across $k$ ranks, each of which holds $N=B/k$ local samples. Each rank all-gathers the local embeddings and
labels, computes logits only with its shard, and uses two reductions to recover
the exact global softmax: an all-reduce of local maxima followed by an
all-reduce of the exponentiated sums. The target probability is obtained by an
all-reduce over the rank owning each target class.

Each rank then forms its local gradient $G_i=(P_i-Y_i)/(Nk)$ and computes
$\nabla W_i=X^TG_i$ and $\nabla X^{(i)}=G_iW_i^T$. Classifier gradients remain
local; a reduce-scatter sums the feature gradients and returns each rank's
original $N$ rows. Thus the distributed procedure is mathematically identical to
full softmax while communicating only embeddings, softmax statistics, target
probabilities, and feature gradients---not the global logits or classifier.

\begin{figure}[tbp]
  \centering
  \includegraphics[width=\textwidth]{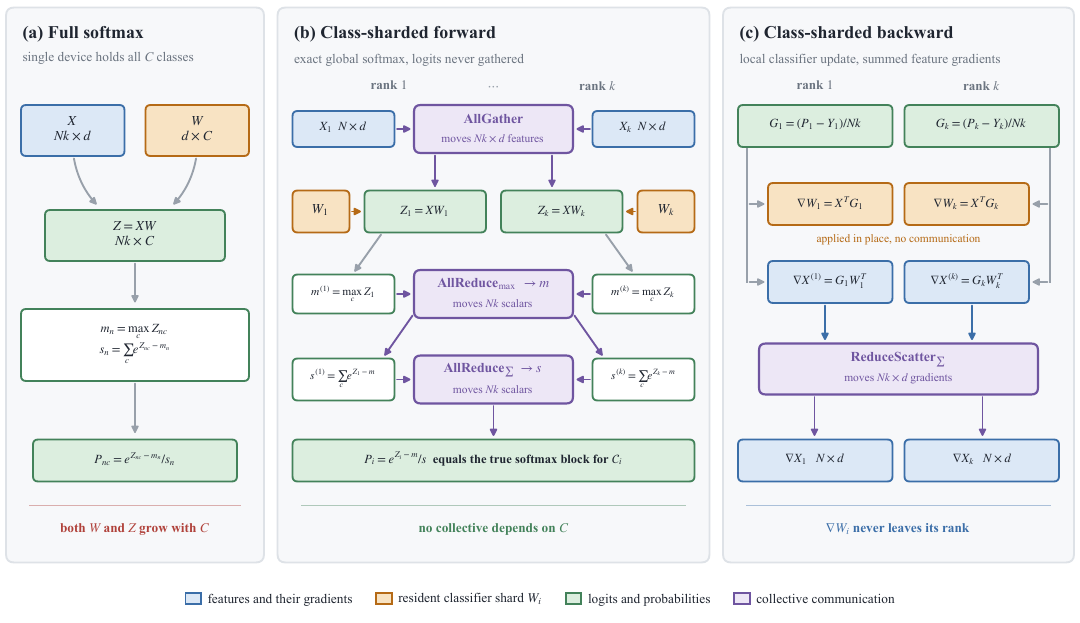}
  \caption{Class-sharded softmax computation. \textbf{(a)}~A single device stores
    the full classifier $W\in\mathbb{R}^{d\times C}$ and materializes all
    $Nk\times C$ logits, so both grow with $C$. \textbf{(b)}~Each rank evaluates
    logits only for its resident shard $W_i\in\mathbb{R}^{d\times C_i}$. An
    all-reduce of local maxima followed by an all-reduce of exponentiated sums
    recovers the exact global normalization, so $P_i$ equals the corresponding
    block of the full softmax without gathering logits. \textbf{(c)}~Classifier
    gradients are computed and applied locally, while reduce-scatter sums the
    feature-gradient contributions and returns each rank its own $N\times d$
    rows. Every collective moves at most $Nk\times d$ values and is independent
    of $C$.}
  \label{fig:model_parallel_explainer}
\end{figure}

\noindent\textbf{Distributed forward pass.}
Figure~\ref{fig:model_parallel_explainer} summarizes the procedure. More
explicitly, rank $i$ initially holds $X_i\in\mathbb{R}^{N\times d}$ and labels
$y_i$. After all-gather, $X=[X_1;\ldots;X_k]\in\mathbb{R}^{Nk\times d}$ is
replicated on every rank, while the classifier remains sharded. Rank $i$
computes $Z_i=XW_i\in\mathbb{R}^{Nk\times C_i}$. It first obtains the local row
maximum $m_n^{(i)}=\max_{c\in\mathcal C_i}Z_{i,nc}$ and computes
$m_n=\operatorname{AllReduce}_{\max}(m_n^{(1)},\ldots,m_n^{(k)})$. It then
evaluates $s_n^{(i)}=\sum_{c\in\mathcal C_i}\exp(Z_{i,nc}-m_n)$ and obtains
$s_n=\operatorname{AllReduce}_{\sum}(s_n^{(1)},\ldots,s_n^{(k)})$. The local
probabilities are therefore $P_{i,nc}=\exp(Z_{i,nc}-m_n)/s_n$, exactly matching
the corresponding block of the global softmax.

\noindent\textbf{Loss and backward pass.} Only the rank owning $y_n$ contributes
its target probability; an all-reduce sum recovers $p_{n,y_n}$ and hence
$L=-(Nk)^{-1}\sum_n\log p_{n,y_n}$. During backpropagation, $G_i=(P_i-Y_i)/(Nk)$
gives the local classifier gradient $\nabla W_i=X^TG_i$. The feature
contributions $\nabla X^{(i)}=G_iW_i^T$ are summed by reduce-scatter, which
returns the gradient for each rank's original local batch.

\subsubsection{Memory Limits of Model Parallelism}
Although sharding reduces the per-GPU classifier storage to $M_W=4dC/k$, every
rank still processes the global batch of size $Nk$. Its local logits therefore
require $M_{\mathrm{logit}}=4NC$ bytes, independent of $k$. With momentum SGD
and margin-based softmax, $M_{\mathrm{FC}}\approx3M_W+2M_{\mathrm{logit}}$.
Thus, when $C/k$ is fixed and more GPUs are added, logit memory grows relative
to the weight storage and becomes the dominant bottleneck. For $N=64$, $d=512$,
and approximately 125K classes per GPU, the logit-to-weight ratio reaches 10 at
80 GPUs, so increasing the GPU count alone cannot make full-class model
parallelism scalable.

\begin{figure}[tbp]
  \centering
  \includegraphics[width=\textwidth]{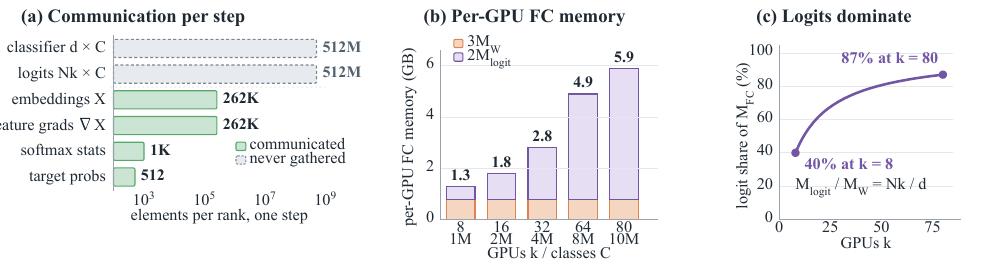}
  \caption{Why exact class-sharded model parallelism still hits a memory wall
  ($N=64$, $d=512$, 125K classes per GPU, so $C$ grows with $k$).
  (a) Elements moved per rank in one step, on a logarithmic axis: only
  embeddings, softmax statistics, target probabilities, and feature gradients
  are communicated, whereas the $Nk\times C$ logits and the $d\times C$
  classifier never leave their owner. (b) Per-GPU FC memory splits into a
  weight term $3M_W$ that is constant under fixed $C/k$ and a logit term
  $2M_{\mathrm{logit}}=8NC$ that grows linearly with $k$. (c) Consequently the
  logit share of $M_{\mathrm{FC}}$, namely
  $2M_{\mathrm{logit}}/(3M_W+2M_{\mathrm{logit}})$, grows with the
  logit-to-weight ratio $M_{\mathrm{logit}}/M_W=Nk/d$ and reaches 87\% at 80
  GPUs.}
  \label{fig:model_parallel_limits}
\end{figure}

Figure~\ref{fig:model_parallel_limits} makes this limit concrete. The procedure
is exact yet communicates almost nothing: one step moves roughly $5\times10^{5}$
elements per rank, about three orders of magnitude less than the $Nk\times C$
logits and the $d\times C$ classifier that are never gathered. The cost
reappears as storage. Because $M_W$ stays constant when $C/k$ is fixed while
$M_{\mathrm{logit}}$ grows linearly with $k$, per-GPU FC memory rises from
$1.3$\,GB at 8 GPUs to $5.9$\,GB at 80 GPUs, and the logit share of
$M_{\mathrm{FC}}$ rises from 40\% to 87\%. Adding ranks therefore buys
classifier capacity but not logit capacity.

These limits arise because model parallelism partitions the classifier but still
evaluates every class. Reducing the class dimension therefore requires an
orthogonal approximation: preserve all positive centers while activating only a
sampled subset of negatives.

\subsection{Approximation Strategy}

\subsubsection{Batch-Level Positive-Preserving Sampling}
Building on the preceding update analysis, PFC constructs one active class set
shared by the mini-batch. Let
$\mathcal{P}=\operatorname{unique}\{y_n\}_{n=1}^{B}$ be the set of positive
classes. Given sampling rate $r$, we sample negative classes $\mathcal{N}$ from
the remaining classifier and form $\mathcal{S}=\mathcal{P}\cup\mathcal{N}$, which
contains at most $rC$ negatives and therefore satisfies $|\mathcal{S}|\approx rC$
whenever $|\mathcal{P}|\le rC$. For every sample $n$, PFC evaluates
\begin{equation}
\widehat p_{n,j}
=\frac{\exp(f_{n,j})}{\sum_{c\in\mathcal{S}}\exp(f_{n,c})}
\qquad j\in\mathcal{S}.
\label{eq:sampled_probability}
\end{equation}
Thus every target class participates in the update, while unsampled negatives
remain unchanged in the current iteration. Because the denominator is restricted
to $\mathcal{S}$, the resulting gradient is not an unbiased full-softmax
gradient. Re-sampling $\mathcal{N}$ across iterations instead provides
continually refreshed negative coverage, with selected hard negatives exerting
larger gradients through their softmax probabilities.

\subsubsection{Distributed Approximation}
\begin{figure}[tbp]
  \centering
  \includegraphics[width=0.94\textwidth]{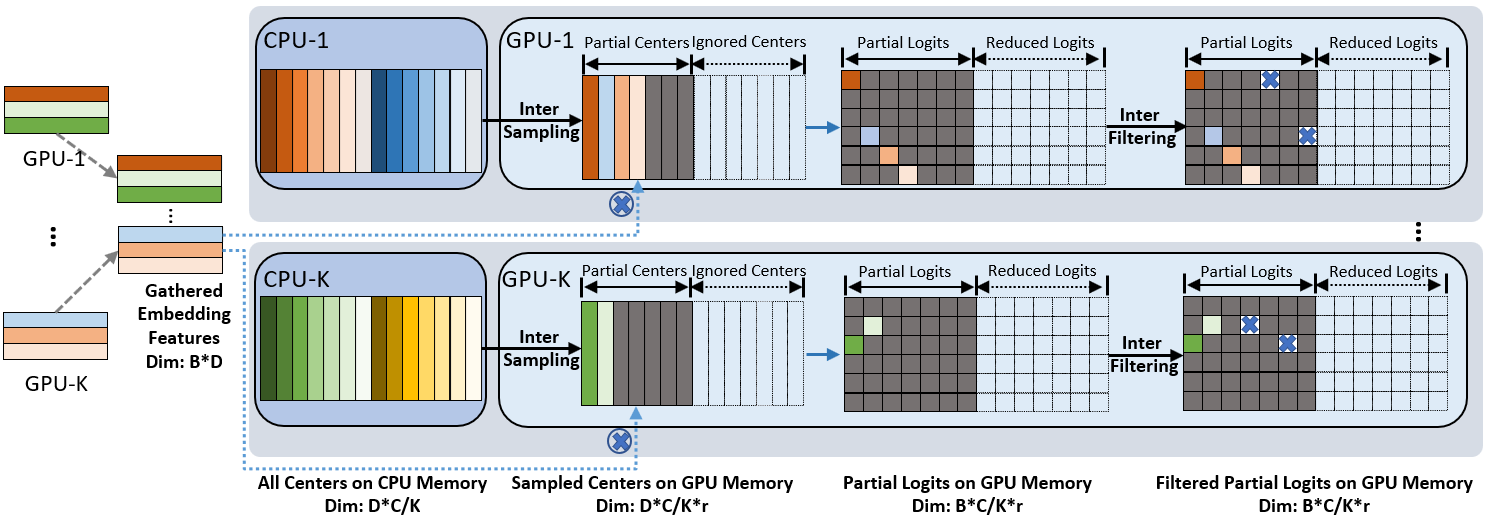}
  \caption{Distributed Partial FC. Features are gathered across GPUs, while each
    rank selects all positive class centers and samples negative centers from
    its owned CPU class shard into a fixed GPU buffer. The resulting partial
    logits reduce classifier memory and computation. Optional filtering removes
    abnormally similar negative centers under severe inter-class conflict.}
  \label{fig:pfc_framework}
\end{figure}
PFC implements the preceding sampling rule with a class-sharded softmax, as
illustrated in Figure~\ref{fig:pfc_framework}. It separates two independent
decisions: \emph{ownership} determines which rank stores a class center, while
\emph{activation} determines whether that center participates in the current
iteration.

For supervised classification, the queries are image embeddings and the
candidates are persistent learnable class centers. PFC combines permanent class
ownership with transient class activation. Let rank $q$ own the contiguous
global class interval
\begin{equation}
\mathcal{C}_q=[a_q,a_q+C_q), \qquad
W_q\in\mathbb{R}^{d\times C_q},
\end{equation}
where $C_q\in\{\lfloor C/k\rfloor,\lceil C/k\rceil\}$ and the intervals
$\{\mathcal{C}_q\}_{q=0}^{k-1}$ form a disjoint partition of all $C$ classes.
Embeddings and labels are gathered across ranks, but the classifier shards are
never gathered.

For every gathered sample $n$ with global label $y_n$, rank $q$ first determines
whether it owns the corresponding center. Labels outside $\mathcal{C}_q$ are
masked by $-1$; owned labels are converted to local coordinates,
\begin{equation}
\tilde y_n^{(q)}=
\begin{cases}
y_n-a_q, & y_n\in\mathcal{C}_q,\\
-1, & \text{otherwise}.
\end{cases}
\end{equation}
The local positive set is therefore
\begin{equation}
\mathcal{P}_q=\operatorname{unique}\{\tilde y_n^{(q)}\mid \tilde y_n^{(q)}\ge 0\}.
\end{equation}

Each rank uses the same relative sampling rate $r$ and a local activation budget
$b_q=\lfloor rC_q\rfloor$. All positive centers are retained. The remaining
budget is filled by sampling without replacement from the local negative pool:
\begin{equation}
\mathcal{S}_q=\mathcal{P}_q\cup
\operatorname{Sample}\!\left(
[0,C_q)\setminus\mathcal{P}_q,
\max(b_q-|\mathcal{P}_q|,0)
\right).
\label{eq:local_sampling}
\end{equation}
If the number of local positives exceeds the nominal budget, all positives are
kept. Sorting $\mathcal{S}_q$ permits an efficient second label mapping via
binary search,
\begin{equation}
\hat y_n^{(q)}=\operatorname{searchsorted}(\mathcal{S}_q,\tilde y_n^{(q)}),
\end{equation}
which maps an owned class to its column in the sampled classifier
$W_q^s=W_q[:,\mathcal{S}_q]$. The union of the sampled global shards forms the
batch-shared active set $\mathcal{S}$ used in Eq.~\ref{eq:sampled_probability}.

The active classifier is thus distributed as $\{W_q^s\}_{q=0}^{k-1}$ rather than
materialized as one global matrix. Each rank computes cosine logits only against
$W_q^s$. Global max and sum reductions produce an exact softmax over the union
of sampled shards, while a reduce-scatter sums the feature gradients and returns
them to their source ranks. Consequently, all positives receive updates, only
sampled negatives are activated, and both classifier memory and computation
scale with $rC_q$ instead of $C_q$. When $r=1$, the procedure reduces to
conventional full-class model parallelism.

\subsubsection{Matrix Multiplication and Communication Efficiency}
\begin{figure}[tbp]
  \centering
  \includegraphics[width=\textwidth]{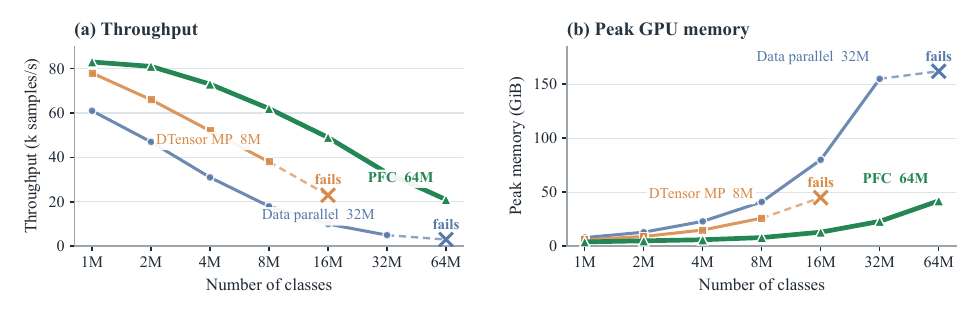}
  \caption{System scaling with the number of classes on eight GPUs, using
    512-dimensional embeddings and full forward--backward execution. The
    backbone is held fixed across all methods, so it contributes a constant cost
    and does not affect the comparison. (a) Training throughput in thousands of
    samples per second. (b) Peak GPU memory in GiB. Data parallel and PFC use
    $r=0.1$, whereas DTensor model parallelism evaluates all classes ($r=1$);
    the direct curve labels follow the corresponding implementations. PFC maintains
    higher throughput and slower memory growth than data parallelism and DTensor
    model parallelism, remaining executable at 64M classes. Crosses denote the
    first out-of-memory or otherwise failed configuration for each baseline.}
  \label{fig:performance_analysis}
\end{figure}

\noindent\textbf{Efficiency and scaling analysis.}
Figure~\ref{fig:performance_analysis} summarizes the system-level scaling
results. We compare PFC against two baselines: data parallelism, which
replicates the sampled classifier on every rank and therefore synchronizes its
gradient, and DTensor model parallelism, which shards the classifier with the
PyTorch DTensor
abstraction\footnote{\url{https://docs.pytorch.org/docs/stable/distributed.tensor.html}}
and evaluates all classes without sampling. All three implementations are
benchmarked under the same PyTorch build, GPUs, embedding dimension, and batch
size, so the reported differences reflect only the classifier design. PFC
maintains substantially higher throughput than data parallelism as the class
count grows, while its memory usage increases much more slowly; the competing
methods fail at progressively smaller classifier sizes.

\noindent\textbf{Matrix multiplication and memory traffic.} Sampling can be
combined with either parallelization strategy. We write DP-PFC for the sampled
data-parallel variant, in which every rank replicates the sampled classifier,
and MP-PFC for the model-parallel variant adopted in this paper, in which each
rank owns a disjoint class shard; full model parallelism denotes the unsampled
case $r=1$. Outside this comparison, PFC always refers to MP-PFC. Sampling is
performed on each owned classifier shard before GEMM. Let every rank receive a
local mini-batch of $N$ embeddings. After all-gather, MP-PFC multiplies
$X\in\mathbb{R}^{Nk\times d}$ by a sampled local classifier
$W_i^s\in\mathbb{R}^{d\times(rC/k)}$:
\begin{align}
\text{DP-PFC:}\quad
&[N\times d][d\times rC],
\label{eq:dp_gemm}\\
\text{MP-PFC:}\quad
&[Nk\times d][d\times rC/k].
\label{eq:mp_gemm}
\end{align}
Both layouts generate $NrC$ logits per rank and have the same leading-order
cost,
\begin{equation}
F_{\mathrm{DP}}=F_{\mathrm{MP}}=2NdrC,
\label{eq:sampled_gemm_flops}
\end{equation}
where one multiply and one addition count as two FLOPs. Their difference is
therefore not explained by arithmetic count alone.

\noindent\textbf{Why identical FLOPs do not imply identical speed.}
A GEMM runs at the tighter of its compute and memory-bandwidth limits. For the
classifier operand, the relevant arithmetic intensity is fixed by the matrix
shape. Every weight element loaded by a rank participates in one
multiply--accumulate for every GEMM row. Consequently, the reuse of a loaded
center is $N$ rows in DP-PFC but $Nk$ rows in MP-PFC. If an operand element
occupies $b$ bytes, their classifier-read intensities are respectively
$2N/b$ and $2Nk/b$ FLOPs per byte. The arithmetic is identical, but MP-PFC
earns $k\times$ more useful work from each byte of classifier data.

\begin{center}
\small
\setlength{\tabcolsep}{8pt}
\textbf{Per-rank classifier geometry.} Reuse counts the GEMM rows served by
each loaded center.\\[3pt]
\begin{tabular}{lcccc}
\toprule
Layout & Centers held & Rows multiplied & Reuse/center & Classifier read \\
\midrule
Full MP    & $C/k$   & $Nk$ & $Nk$ & $Cd/k$ \\
DP-PFC     & $rC$    & $N$  & $N$  & $rCd$ \\
MP-PFC     & $rC/k$  & $Nk$ & $Nk$ & $rCd/k$ \\
\bottomrule
\end{tabular}
\end{center}

\begin{figure}[tbp]
\centering
\includegraphics[width=\textwidth]{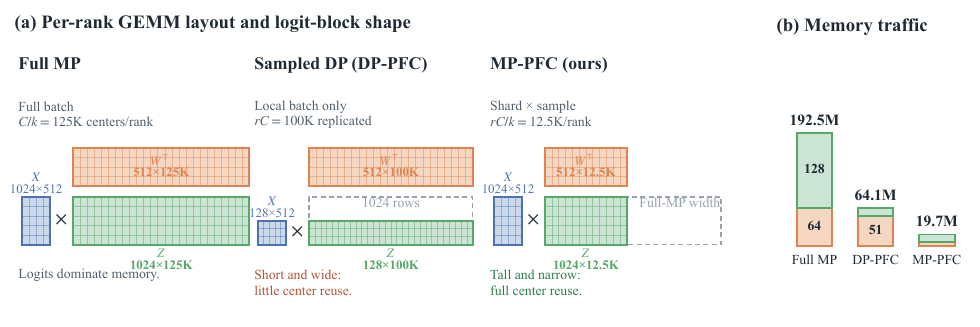}
\captionsetup{font={small,stretch=1.12}}
\caption{Per-rank GEMM layout and memory traffic ($k=8$, $N=128$, $C=10^6$,
  $r=0.1$, $d=512$). (a) Per-rank GEMM shapes. Block edges are drawn as
  $(\text{dimension})^{1/3}$ so that the true dimensions of $X$, $W^\top$, and
  $Z$ remain legible despite the much larger class axis. The dashed box in
  DP-PFC marks the rows that a gathered batch would have filled; the dashed box
  in MP-PFC marks the full-MP class width that sampling removes. Thus DP-PFC is
  short and wide, while MP-PFC keeps the gathered-batch rows and shrinks the
  class axis to $rC/k$. (b) First-order traffic per rank, stacked as feature
  read, weight read, and logit write. MP-PFC reduces the total to $19.7$M
  elements against $64.1$M for DP-PFC and $192.5$M for full model parallelism,
  and cuts the weight read alone by $10\times$ relative to full model
  parallelism.}
\label{fig:gemm_memory_traffic}
\end{figure}

This is why the short, wide block in
Figure~\ref{fig:gemm_memory_traffic}(a) is unfavorable. Its small row count
provides little opportunity to amortize each weight load, so classifier
streaming can become the binding roofline. MP-PFC performs the same
$2NdrC$ FLOPs with a $k\times$ smaller local weight operand and $k\times$
greater reuse per loaded center. In the illustrated setting, its
$1024\times12.5$K logit block has a class-to-row aspect ratio of about
$12{:}1$, against $122{:}1$ for full MP and $781{:}1$ for DP-PFC.

Ignoring cache effects and subsequent softmax accesses, the compulsory
first-order traffic per rank is
\begin{align}
T_{\mathrm{DP}} &\approx Nd+rCd+NrC,\\
T_{\mathrm{MP}} &\approx Nkd+\frac{rCd}{k}+NrC.
\label{eq:gemm_traffic}
\end{align}
Sampling and sharding therefore reduce local classifier traffic
multiplicatively, from $Cd$ to $rCd/k$. Under the
Figure~\ref{fig:gemm_memory_traffic} setting, MP-PFC moves $19.7$M elements per
rank, compared with $64.1$M for DP-PFC and $192.5$M for full model parallelism.

The reduction can also cross a cache-capacity threshold. GEMM kernels revisit
weight tiles as they advance along the row dimension. Whether a revisit is
served from cache or HBM depends on the effective operand footprint, data type,
tiling policy, and concurrent cache use. Sampling and sharding reduce that
footprint by $r/k$ relative to the full classifier; once the active shard fits
in cache, the cost of row-wise reuse can change discontinuously rather than by
only the nominal traffic ratio. We therefore treat cache residency as a second
shape-dependent benefit, not as part of the compulsory-traffic estimate above.

Finally, Figure~\ref{fig:gemm_memory_traffic}(b) counts only the forward GEMM's
feature read, weight read, and logit write. The logit block is subsequently
read for softmax normalization and again during backward. It is the only term
that scales with the product of batch and class count, $NrC$, and sampling
shrinks it by $1/r$ relative to unsampled full model parallelism. Thus the
weight-shard argument explains why the two sampled GEMMs can run at different
speeds, while the logit argument explains why full model parallelism remains
expensive across the complete classifier pipeline.

\noindent\textbf{Communication volume.} The communication patterns differ more
fundamentally. In DP-PFC, every rank maintains a replica of the sampled
classifier. Its sampled weight gradient must be synchronized after
backpropagation. Omitting collective-algorithm constants, the communicated
payload per iteration is proportional to
\begin{equation}
V_{\mathrm{DP}}=\Theta(rCd),
\label{eq:dp_comm}
\end{equation}
which remains large when $C$ is in the millions, even after sampling. This
estimate assumes that ranks use a compatible sampled class set; independent
samples require sparse index exchange or an equivalent synchronization
mechanism.

In MP-PFC, each class center has a single owner, so sampled weight gradients
remain local and require no classifier-gradient all-reduce. Communication
instead consists of gathering embeddings before the classifier, returning their
accumulated gradients during backward, and reducing a few softmax statistics per
sample:
\begin{equation}
V_{\mathrm{MP}}=\Theta(Nkd)+\Theta(Nk).
\label{eq:mp_comm}
\end{equation}
The first term represents embedding and feature-gradient communication; the
second represents global maximum, denominator, and target-probability
reductions. Crucially, this cost is independent of $C$ and nearly independent of
$r$. Thus sampling and model parallelism are complementary: sampling reduces
local GEMM and logit storage, while class ownership removes the $\Theta(rCd)$
classifier-gradient synchronization that remains in DP-PFC.

\begin{figure}[tbp]
\centering
\includegraphics[width=\textwidth]{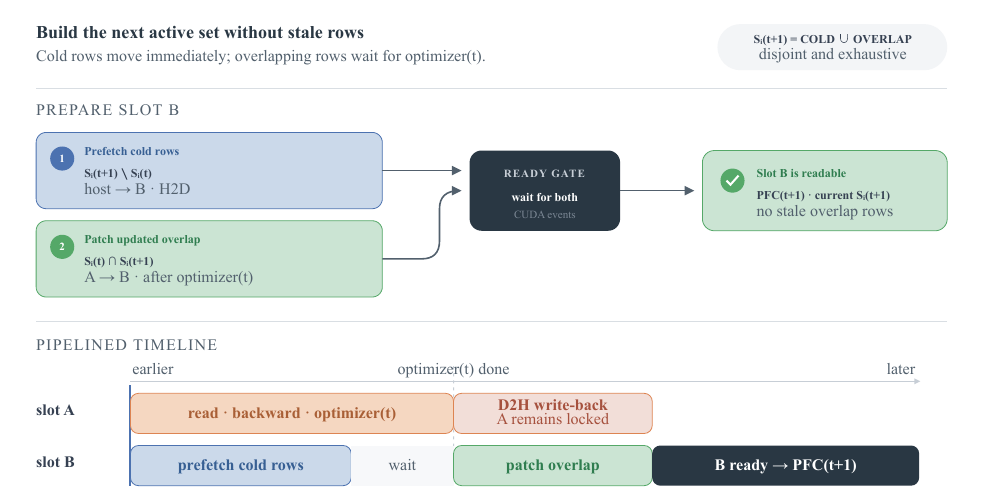}
\caption{Stale-free construction of the next active classifier set. Cold rows
  are prefetched into slot B while PFC$(t)$ uses slot A; rows shared by
  iterations $t$ and $t+1$ are patched only after optimizer$(t)$. B becomes
  readable only after both paths complete, and A becomes reusable only after
  its readers and D2H write-back complete. Row versions validate host state,
  while CUDA events enforce the two readiness invariants.}
\label{fig:async_pipeline}
\end{figure}

\noindent\textbf{Asynchronous hierarchical class storage.} Partial activation
also allows the complete class bank and optimizer state to reside outside GPU
memory. Unsampled centers can remain in pinned host memory or an NVMe-backed
cache, while only the rows in $\mathcal{S}_q$ are staged on the GPU. Labels for
mini-batch $t+1$ can determine the next active rows during iteration $t$; those
rows are prefetched on a dedicated CUDA stream and overlapped with backbone
computation. The next set is split into cold rows
$\mathcal{S}_i(t+1)\setminus\mathcal{S}_i(t)$ and overlapping rows
$\mathcal{S}_i(t)\cap\mathcal{S}_i(t+1)$. Cold rows may be prefetched into slot
B immediately, whereas overlapping rows are copied from slot A only after
optimizer step $t$ has produced their current values. A CUDA-event ready gate
exposes B to PFC$(t+1)$ only after both paths finish. After classifier backward,
sampled gradients and updated rows are written back asynchronously; slot A
remains locked until its readers and D2H write-back complete. Per-row versioning
validates host state, while double-buffered slots and events enforce device-side
ordering. Figure~\ref{fig:async_pipeline} illustrates these dependencies. This
reduces GPU-resident classifier storage from $\Theta(C_q d)$ to $\Theta(rC_q
d)$, provided that data movement fits within the computation window.

Beyond systems efficiency, partial activation also changes how often potentially
conflicting negatives affect optimization. We next exploit this property to
improve robustness to noisy web-scale labels.

\subsubsection{Noise-Robust Inter-Class Filtering}
Web-scale face datasets inevitably contain inter-class conflicts: images of one
identity may be assigned to multiple class labels. Under full softmax, every
conflicted center participates as a negative whenever its duplicate identity
appears in the mini-batch. The resulting gradient simultaneously pulls the
embedding toward its labeled center and pushes it away from a semantically
positive but nominally negative center. Repeated updates force the classifier to
separate duplicate classes and can overfit annotation noise.

PFC reduces this failure mode before any explicit filtering. If a conflicted
negative center is not a positive center in the current mini-batch, it is
activated with probability approximately $r$. Its expected harmful contribution
to the feature gradient therefore decreases from every iteration under full FC
to only an $r$ fraction of iterations under PFC. Genuine hard negatives are
still sampled and optimized over time, while conflicted negatives receive fewer
contradictory updates.

For heavily corrupted data, we further use the PFC* variant of Partial
FC~\cite{an2022killing}. Filtering is applied independently to each
sample--class pair, based on the unmargined cosine similarity. For sample $n$
and active class $j\in\mathcal{S}$, define
\begin{equation}
M_{n,j}=\mathbf{1}[j=y_n]
+\mathbf{1}[j\ne y_n]\,\mathbf{1}[\cos\theta_{n,j}\leq\tau],
\qquad \tau=0.4.
\label{eq:noise_filtering}
\end{equation}
Only logits with $M_{n,j}=1$ enter sample $n$'s denominator. Positive centers
are never filtered, while an active center may be suppressed as a negative for
one sample and retained for another. The rule targets only abnormally similar
negatives, which are more likely to represent duplicate identities than useful
inter-class boundaries. Unlike global pair cleaning, it operates online on the
already sampled logits and adds no full-class search. The threshold should
nevertheless be enabled only when substantial label conflict is expected,
because aggressive filtering may remove legitimate hard negatives. We write
FC*-1.0 for the same filter applied to full softmax, that is, PFC* with $r=1$.

In summary, PFC combines positive-preserving sampling, permanent class
ownership, and optional conflict filtering. The following experiments evaluate
these three aspects through system scaling, recognition accuracy, and robustness
under imperfect training data.

\clearpage
\section{Experiments}
\subsection{Datasets and Settings}
\noindent\textbf{Training Dataset.} Our training datasets include CASIA
\cite{yi2014learning}, VGGFace \cite{parkhi2015deep}, GlintAsian, MegaFaceMS1M
\cite{kemelmacher-shlizerman2016the}, MS1MV2 \cite{deng2019arcface}, and MS1MV3
\cite{deng2019lightweight}; dataset names follow the InsightFace Dataset Zoo. We
also clean Celeb-500K \cite{cao2018celeb} and
merge it with MS1MV2 to construct \textbf{Glint360K}, containing 17 million
images from 360K identities. For the extended experiments, we use the WebFace
family, including WebFace4M, WebFace12M, and WebFace42M, which contain
approximately 4M, 12M, and 42M training images, respectively. The WebFace
experiments additionally use WebFace12M-Conflict to study inter-class conflicts,
label noise, and long-tailed identity distributions.

\noindent\textbf{Evaluation Benchmarks.} Following the evaluation protocol used
in ArcFace and Partial FC, we evaluate face verification on standard benchmarks,
including LFW
\cite{huang2008labeled}, CFP-FP \cite{sengupta2016frontal}, and AgeDB-30
\cite{moschoglou2017agedb}. We further report large-scale verification
results on IJB-C \cite{maze2018iarpa} and the InsightFace
ICCV21-MFR\footnote{\url{https://github.com/deepinsight/insightface/tree/master/IFRT}}.

\noindent\textbf{Training Settings.} We use IR50 and IR100 as backbones, that
is, the improved-ResNet variants of depth 50 and 100 introduced for face
recognition in ArcFace, in which the standard ResNet stem and residual unit are
adapted to low-resolution aligned face crops~\cite{deng2019arcface,he2016deep}.
Both are trained with CosFace or ArcFace. The feature scale is set to $s=64$.
The cosine margin for CosFace is $0.4$, and the angular margin for ArcFace is
$0.5$. For the original CASIA, MS1MV2, and Glint360K experiments, the total
batch size is 512 across eight NVIDIA RTX2080Ti GPUs, and the initial learning
rate is $0.1$. For CASIA, the learning rate is divided by 10 at 20K and 28K
iterations, and training ends at 32K iterations. For MS1MV2, it is divided by 10
at 100K and 160K iterations, and training ends at 180K iterations. For
Glint360K, it is divided by 10 at 200K, 400K, 500K, and 550K iterations, and
training ends at 600K iterations. The WebFace experiments instead follow the
official ArcFace-Torch configurations described in
Section~\ref{sec:extended_experiments}. In particular, WebFace42M IR backbones
use 32 GPUs, a per-GPU batch size of 128, SGD with an initial learning rate of
$0.4$, 20 training epochs, and two warm-up epochs.

\subsection{Benchmark Results}
We report the main face-recognition results in Tables~\ref{tab:mfr}
and~\ref{tab:comprehensive_verification}. Metrics and operating points are
specified in the table captions.
\begin{table}[H]
  \centering
  \footnotesize
  \caption{Face-verification performance across training datasets, using an IR50
    backbone. All result columns are reported on a percentage scale. ICCV21-MFR
    and IJB-C report TAR (\%) at FAR $=10^{-6}$ and $10^{-4}$, respectively;
    LFW, CFP-FP, and AgeDB report verification accuracy (\%). Each baseline/PFC
    pair uses the same backbone, margin loss, data, and training schedule;
    ``+PFC'' changes only the classifier to Partial FC with sampling rate
    $r=0.1$. Boldface marks the higher ICCV21-MFR All result within each pair.
    Dataset names follow the
    \href{https://github.com/deepinsight/insightface/wiki/Dataset-Zoo}{InsightFace
    Dataset Zoo}.}
  \label{tab:mfr}
  \setlength{\tabcolsep}{3.8pt}
  \renewcommand{\arraystretch}{1.08}
  \resizebox{\textwidth}{!}{%
  \begin{tabular}{@{}>{\raggedright\arraybackslash}p{0.24\textwidth}*{9}{>{\centering\arraybackslash}p{0.084\textwidth}}@{}}
    \toprule
    \multirow{2}{*}{\textbf{Training dataset}} & \multicolumn{5}{c}{\textbf{ICCV21-MFR}} & \multicolumn{4}{c}{\textbf{Standard benchmarks}} \\
    \cmidrule(lr){2-6}\cmidrule(lr){7-10}
    & \textbf{All} & \textbf{African} & \makecell{\textbf{Cauca-}\\\textbf{sian}} & \makecell{\textbf{South}\\\textbf{Asian}} & \makecell{\textbf{East}\\\textbf{Asian}} & \textbf{LFW} & \textbf{CFP-FP} & \textbf{AgeDB} & \textbf{IJB-C} \\
    \midrule
    CASIA~\cite{yi2014learning}             & 36.8 & 42.6 & 55.8 & 49.6 & 19.6 & 99.5 & 95.2 & 94.9 & 87.2 \\
    \textbf{CASIA + PFC}       & \textbf{37.1} & 38.9 & 53.8 & 48.7 & 19.9 & 99.4 & 95.4 & 94.6 & 85.0 \\
    \addlinespace[2pt]
    VGGFace~\cite{parkhi2015deep}           & 38.6 & 35.3 & 54.3 & 44.1 & 24.1 & 99.6 & 97.4 & 95.1 & 91.2 \\
    \textbf{VGGFace + PFC}     & \textbf{40.7} & 36.8 & 60.2 & 49.0 & 24.3 & 99.7 & 98.5 & 95.4 & 92.5 \\
    \addlinespace[2pt]
    GlintAsian        & 62.7 & 49.5 & 64.8 & 58.0 & 61.7 & 99.6 & 93.2 & 95.4 & 91.5 \\
    \textbf{GlintAsian + PFC}  & \textbf{63.2} & 50.4 & 65.2 & 57.9 & 61.8 & 99.7 & 93.0 & 95.2 & 91.1 \\
    \addlinespace[2pt]
    MS1MV2~\cite{deng2019arcface}            & 77.7 & 74.6 & 84.1 & 82.0 & 51.1 & 99.8 & 98.1 & 98.1 & 96.1 \\
    \textbf{MS1MV2 + PFC}      & \textbf{77.7} & 74.7 & 84.9 & 82.8 & 52.5 & 99.8 & 98.1 & 98.0 & 96.1 \\
    \addlinespace[2pt]
    MegaFaceMS1M~\cite{kemelmacher-shlizerman2016the}      & 78.4 & 74.1 & 82.3 & 77.2 & 60.2 & 99.8 & 97.6 & 97.4 & 95.4 \\
    \textbf{MegaFaceMS1M + PFC}& \textbf{78.8} & 73.7 & 83.0 & 78.8 & 57.6 & 99.8 & 97.9 & 97.7 & 95.4 \\
    \addlinespace[2pt]
    MS1MV3~\cite{deng2019lightweight}            & \textbf{82.5} & 77.2 & 87.0 & 86.0 & 60.6 & 99.8 & 98.5 & 98.3 & 96.6 \\
    \textbf{MS1MV3 + PFC}      & 81.7 & 78.1 & 87.3 & 85.5 & 58.9 & 99.8 & 98.4 & 98.2 & 96.4 \\
    \addlinespace[2pt]
    Glint360K~\cite{an2022killing}         & 86.8 & 84.8 & 91.4 & 90.1 & 66.2 & 99.8 & 99.1 & 98.5 & 97.1 \\
    \textbf{Glint360K + PFC}   & \textbf{87.1} & 85.3 & 91.6 & 90.5 & 66.8 & 99.8 & 99.1 & 98.5 & 97.0 \\
    \addlinespace[2pt]
    WebFace12M~\cite{zhu2021webface260m}        & \textbf{90.6} & 89.4 & 94.2 & 92.4 & 73.9 & 99.8 & 99.2 & 98.1 & 97.1 \\
    \textbf{WebFace12M + PFC}  & 90.0 & 89.3 & 94.0 & 92.4 & 73.0 & 99.8 & 99.1 & 98.1 & 97.0 \\
    \bottomrule
  \end{tabular}
  }
\end{table}
\begin{table}[tbp]
  \centering
  \footnotesize
  \caption{Face-verification performance across methods and sampling rates. All
    result columns are reported on a percentage scale. All PFC models use an
    IR100 backbone. IJB-B and IJB-C report TAR (\%) at FAR $=10^{-4}$. MegaFace
    Challenge 1 uses FaceScrub as the probe set: Id is rank-1 identification
    (\%) with one million distractors, and Ver is verification TAR (\%) at FAR
    $=10^{-6}$. The remaining columns report verification accuracy (\%). The
    sampling rate $r=1.0$ is equivalent to full FC, whereas $r=0.1$ retains
    every positive class and activates 10\% of classes in total.}
  \label{tab:comprehensive_verification}
  \setlength{\tabcolsep}{3.8pt}
  \renewcommand{\arraystretch}{1.08}
  \resizebox{\textwidth}{!}{%
  \begin{tabular}{@{}l*{9}{c}@{}}
    \toprule
    \multirow{2}{*}{\textbf{Method}} & \multicolumn{2}{c}{\textbf{IJB}} & \multicolumn{2}{c}{\textbf{MegaFace}} & \multicolumn{5}{c}{\textbf{Verification accuracy}} \\
    \cmidrule(lr){2-3}\cmidrule(lr){4-5}\cmidrule(lr){6-10}
    & \textbf{IJB-B}~\cite{whitelam2017iarpa} & \textbf{IJB-C}~\cite{maze2018iarpa} & \textbf{Id} & \textbf{Ver} & \textbf{LFW} & \textbf{AgeDB} & \textbf{CALFW}~\cite{zheng2017cross} & \textbf{CPLFW}~\cite{CPLFWTech} & \textbf{CFP-FP} \\
    \midrule
    CosFace ($m=0.35$)~\cite{wang2018cosface} & -- & -- & 97.9 & 97.9 & 99.4 & -- & 90.6 & 84.0 & -- \\
    ArcFace ($m=0.5$)~\cite{deng2019arcface} & 94.2 & 95.6 & 98.4 & 98.5 & 99.8 & -- & 95.5 & 92.1 & 98.3 \\
    GroupFace~\cite{kim2020groupface} & 94.9 & 96.3 & 98.7 & 98.8 & 99.9 & 98.3 & 96.2 & 93.2 & 98.6 \\
    Circle Loss~\cite{sun2020circle} & -- & 94.0 & 98.5 & 98.7 & 99.7 & -- & -- & -- & -- \\
    CurricularFace~\cite{huang2020curricularface} & 94.8 & 96.1 & 98.7 & 98.6 & 99.8 & 98.3 & 96.2 & 93.1 & 98.4 \\
    \midrule
    MS1MV2, CosFace, PFC ($r=1.0$) & 95.0 & 96.4 & 98.4 & 98.6 & 99.8 & 98.0 & 96.2 & 93.1 & 98.5 \\
    \textbf{MS1MV2, CosFace, PFC ($r=0.1$)} & 94.6 & 96.0 & 98.0 & 98.5 & 99.8 & 98.1 & 96.1 & 92.9 & 98.6 \\
    \addlinespace[2pt]
    MS1MV2, ArcFace, PFC ($r=1.0$) & 94.8 & 96.2 & 98.3 & 98.6 & 99.8 & 98.2 & 96.2 & 93.0 & 98.5 \\
    \textbf{MS1MV2, ArcFace, PFC ($r=0.1$)} & 94.4 & 95.8 & 98.3 & 98.0 & 99.8 & 98.2 & 96.2 & 93.0 & 98.5 \\
    \midrule
    Glint360K, CosFace, PFC ($r=1.0$) & 96.1 & 97.3 & 99.1 & 99.0 & 99.8 & 98.6 & 96.2 & 94.8 & 99.3 \\
    \textbf{Glint360K, CosFace, PFC ($r=0.1$)} & 96.1 & 97.2 & 98.9 & 99.1 & 99.8 & 98.6 & 96.2 & 94.8 & 99.3 \\
    \bottomrule
  \end{tabular}
  }
\end{table}
\par\noindent\textbf{Results on IJB-C (Table~\ref{tab:mfr}).}
IJB-C~\cite{maze2018iarpa} contains 3,531 subjects, 31.3K still images, and
117.5K frames from 11,779 videos. Under the 1:1 verification protocol, it
comprises 23,124 templates, 19,557 genuine pairs, and 15.6M impostor pairs. We
follow the ArcFace template-based evaluation procedure: an embedding is
extracted for every detected and aligned face, and the face-detection score and
feature norm are used as quality cues to weight image- and frame-level features
when constructing each template representation. Verification scores are then
computed between template representations. Table~\ref{tab:mfr} reports TAR at
FAR $=10^{-4}$.

\par\noindent\textbf{Results on ICCV21-MFR (Table~\ref{tab:mfr}).} ICCV21-MFR is
the multi-racial evaluation set of the ICCV21 Masked Face Recognition
Challenge~\cite{deng2021masked}. It contains 242,143 identities and 1,624,305
web images. We report overall and group-wise TAR at FAR $=10^{-6}$.

\par\noindent\textbf{Results on MegaFace
(Table~\ref{tab:comprehensive_verification}).} MegaFace Challenge 1 evaluates
face recognition at million-scale distraction using FaceScrub as the probe set
and one million distractor images~\cite{kemelmacher-shlizerman2016the}. We
report rank-1 identification accuracy (Id) and verification TAR (Ver) at FAR
$=10^{-6}$. These metrics test whether PFC preserves both closed-set retrieval
and open-set verification performance as the gallery grows to one million
distractors.

\par\noindent\textbf{Comprehensive comparison
(Table~\ref{tab:comprehensive_verification}).} Across IJB-B, IJB-C, MegaFace,
and five standard verification benchmarks, sampling only 10\% of the classes
largely preserves full-FC accuracy for both CosFace and ArcFace. On MS1MV2, the
differences between $r=0.1$ and $r=1.0$ are at most 0.6 percentage points, while
several results are unchanged or slightly improved. With the larger Glint360K
training set, PFC at $r=0.1$ matches full FC on six of the nine reported
metrics and remains within 0.2 points on all others, while achieving results
competitive with the listed state-of-the-art methods. These results show that
most classifier classes can be omitted at each iteration without materially
degrading verification or identification accuracy.

\subsection{Communication Profiling}
\label{sec:communication_profiling}
\begin{figure}[tbp]
  \centering
  \includegraphics[width=\textwidth]{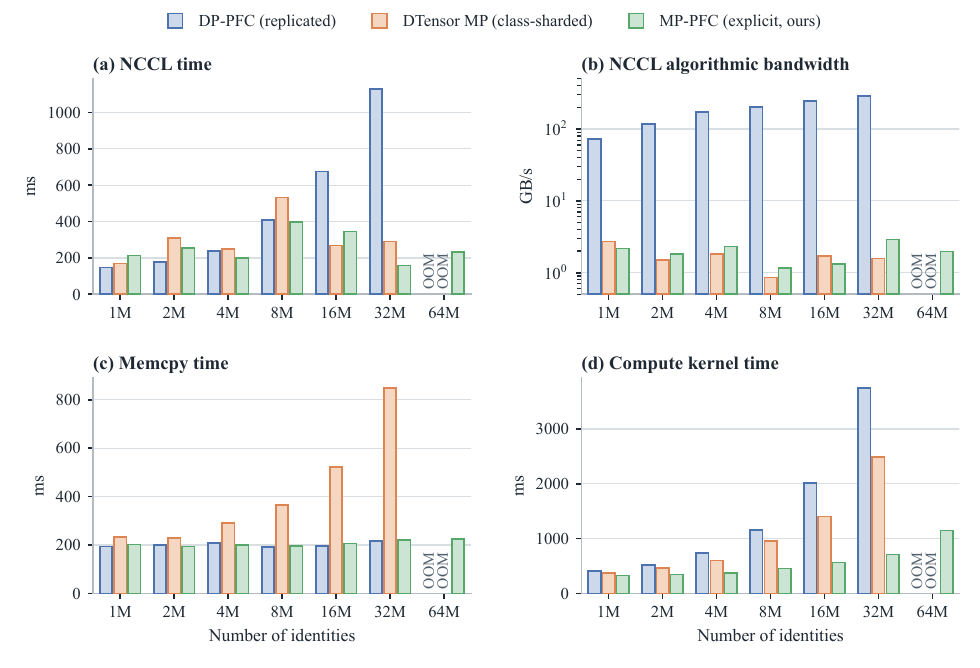}
  \caption{Per-step profile of the three classifier implementations on
    8$\times$H800 at global batch size 2048, as a function of the number of
    identities. Panels (a), (c) and (d) report NCCL, memcpy, and compute-kernel
    time; panel (b) reports the algorithmic NCCL bandwidth on a logarithmic
    axis. As noted in Table~\ref{tab:comm_profile}, the bandwidth in panel (b)
    is computed from profiler-visible bytes only, which for the DTensor and
    MP-PFC bars excludes the class-sharded classifier traffic; the gap in that
    panel therefore reflects byte visibility, not link utilisation. Bars marked
    ``OOM'' did not complete.}
  \label{fig:pfc_comm_profile}
\end{figure}

Accuracy parity settles what PFC costs statistically; it does not say where the
wall-clock time goes. We therefore profile a single training step of the three
classifier implementations with the PyTorch profiler and aggregate the Chrome
trace by event category. The setting is one node of 8$\times$H800, global batch
size 2048 (256 per GPU), a ViT-S/16 backbone at $112\times112$ with embedding
size 512, BF16 autocast, AdamW, and a synthetic data iterator, so that the
measurement isolates the classifier from input-pipeline noise. The three
implementations are the replicated sampled classifier (DP-PFC, $r=0.1$), a
class-sharded DTensor classifier at $r=1.0$, and our explicit rank-local
Partial FC (MP-PFC, $r=0.1$). Table~\ref{tab:comm_profile} lists the full
per-category breakdown and Figure~\ref{fig:pfc_comm_profile} plots it.

\begin{table}[tbp]
  \centering
  \footnotesize
  \caption{Per-step communication and compute profile of the three classifier
    implementations on 8$\times$H800 at global batch size 2048, aggregated from
    PyTorch profiler Chrome traces. ``Ovl.'' is the fraction of the category's
    duration that overlaps a compute kernel. ``OOM'' marks configurations that
    did not complete. Bandwidth columns are algorithmic estimates, defined as
    profiler-visible tensor bytes divided by event duration, not hardware
    counter measurements. \emph{They are not comparable across implementations}:
    the visible NCCL byte count for the DTensor and MP-PFC rows is constant at
    $4.6\times10^{8}$~B at every identity count, i.e.\ it captures only the
    backbone gradient all-reduce and misses the class-sharded classifier
    traffic, so those GB/s values are lower bounds. The time columns are
    directly comparable.}
  \label{tab:comm_profile}
  \setlength{\tabcolsep}{6.5pt}
  \renewcommand{\arraystretch}{1.08}
  \begin{tabular}{@{}ll*{3}{c}*{3}{c}c@{}}
    \toprule
    \multirow{2}{*}{\textbf{Identities}} & \multirow{2}{*}{\textbf{Classifier}} & \multicolumn{3}{c}{\textbf{NCCL}} & \multicolumn{3}{c}{\textbf{Memcpy}} & \multirow{2}{*}{\makecell{\textbf{Kernel}\\\textbf{(ms)}}} \\
    \cmidrule(lr){3-5}\cmidrule(lr){6-8}
    & & \textbf{ms} & \textbf{GB/s} & \textbf{Ovl.} & \textbf{ms} & \textbf{GB/s} & \textbf{Ovl.} & \\
    \midrule
    \multirow{3}{*}{1M}  & DP-PFC  & 147.1 & 72.6  & 69.3\% & 193.6 & 6.5   & 42.5\% & 406.1 \\
                         & DTensor & 169.1 & 2.7   & 23.3\% & 233.8 & 25.1  & 43.8\% & 373.7 \\
                         & \textbf{MP-PFC}  & 213.0 & 2.2   & 27.7\% & 202.5 & 3.7   & 42.8\% & \textbf{326.6} \\
    \addlinespace[2pt]
    \multirow{3}{*}{2M}  & DP-PFC  & 178.9 & 116.9 & 73.5\% & 201.1 & 8.9   & 43.5\% & 520.5 \\
                         & DTensor & 308.4 & 1.5   & 61.0\% & 228.7 & 48.1  & 47.8\% & 468.7 \\
                         & \textbf{MP-PFC}  & 253.4 & 1.8   & 43.8\% & 194.2 & 3.9   & 42.6\% & \textbf{345.2} \\
    \addlinespace[2pt]
    \multirow{3}{*}{4M}  & DP-PFC  & 239.6 & 172.7 & 80.6\% & 209.5 & 13.6  & 41.9\% & 734.2 \\
                         & DTensor & 250.7 & 1.8   & 53.7\% & 292.0 & 72.7  & 50.2\% & 603.0 \\
                         & \textbf{MP-PFC}  & 199.2 & 2.3   & 43.2\% & 200.6 & 3.8   & 41.7\% & \textbf{367.7} \\
    \addlinespace[2pt]
    \multirow{3}{*}{8M}  & DP-PFC  & 409.6 & 201.1 & 92.9\% & 193.1 & 25.6  & 42.5\% & 1161.2 \\
                         & DTensor & 532.0 & 0.9   & 74.9\% & 366.2 & 113.9 & 55.1\% & 960.9 \\
                         & \textbf{MP-PFC}  & 397.8 & 1.2   & 49.0\% & 197.1 & 3.9   & 42.7\% & \textbf{458.5} \\
    \addlinespace[2pt]
    \multirow{3}{*}{16M} & DP-PFC  & 675.1 & 243.3 & 94.0\% & 196.9 & 46.5  & 41.7\% & 2012.6 \\
                         & DTensor & 268.6 & 1.7   & 51.8\% & 521.2 & 158.6 & 60.1\% & 1402.9 \\
                         & \textbf{MP-PFC}  & 346.2 & 1.3   & 56.0\% & 206.0 & 3.8   & 43.9\% & \textbf{562.4} \\
    \addlinespace[2pt]
    \multirow{3}{*}{32M} & DP-PFC  & 1129.0 & 290.6 & 94.2\% & 218.1 & 80.5  & 40.5\% & 3741.0 \\
                         & DTensor & 291.0 & 1.6   & 54.3\% & 847.5 & 194.2 & 64.7\% & 2491.6 \\
                         & \textbf{MP-PFC}  & 158.3 & 2.9   & 29.5\% & 221.9 & 3.6   & 43.5\% & \textbf{713.4} \\
    \addlinespace[2pt]
    \multirow{3}{*}{64M} & DP-PFC  & \multicolumn{7}{c}{OOM} \\
                         & DTensor & \multicolumn{7}{c}{OOM} \\
                         & \textbf{MP-PFC}  & 233.7 & 2.0   & 46.4\% & 223.9 & 3.8   & 47.0\% & \textbf{1146.4} \\
    \addlinespace[2pt]
    \multirow{3}{*}{128M}& DP-PFC  & \multicolumn{7}{c}{OOM} \\
                         & DTensor & \multicolumn{7}{c}{OOM} \\
                         & MP-PFC  & \multicolumn{7}{c}{OOM} \\
    \bottomrule
  \end{tabular}
\end{table}

\par\noindent\textbf{Compute time is where the classifier is decided.} The
kernel column separates the three implementations far more cleanly than any
communication column. At 32M identities MP-PFC spends 713~ms in compute kernels
against 2492~ms for the DTensor classifier and 3741~ms for DP-PFC, a
$3.5\times$ and $5.2\times$ reduction respectively, and the gap widens
monotonically with the identity count. This is the behaviour predicted by the
per-rank GEMM analysis in Figure~\ref{fig:gemm_memory_traffic}: sharding the
sampled class set leaves each rank with an $rC/k$ weight block, so the
classifier GEMM grows $k$ times more slowly than under either alternative.

\par\noindent\textbf{Memcpy, not NCCL, is what breaks the DTensor path.} The
DTensor memcpy time grows from 234~ms to 847~ms between 1M and 32M identities
and becomes the second-largest cost in the step, whereas MP-PFC stays flat in
the 194--224~ms band across the entire sweep, including at 64M. The
class-sharded DTensor classifier materialises and redistributes its sharded
logits through device-to-device copies whose volume scales with $C$; the
explicit implementation keeps the sampled block rank-local and never pays that
term. NCCL time, by contrast, is non-monotonic and comparatively small for both
model-parallel variants.

\par\noindent\textbf{Memory capacity is the binding constraint.} DP-PFC and the
DTensor classifier both fail at 64M identities, while MP-PFC completes and only
fails at 128M. Since all three share a backbone, an optimizer, and a batch size,
the difference is entirely the classifier's resident state: $rC$ replicated
centers for DP-PFC and $C/k$ for the DTensor variant, against $rC/k$ for
MP-PFC. The profiling sweep therefore reproduces, at the level of a single
measured step, the scaling argument made analytically earlier in the paper.

\FloatBarrier
\subsection{Ablation Studies on WebFace}
\label{sec:extended_experiments}

These experiments follow the evaluation protocol of Partial
FC~\cite{an2022killing}. WebFace4M, WebFace12M, and WebFace42M contain 200K,
600K, and 2M identities, respectively. IR backbones are trained for 20 epochs
with SGD and a per-GPU batch size of 128 unless stated otherwise.

\noindent\textbf{Dataset characteristics.} WebFace42M contains both inter-class
conflicts and a pronounced long tail: 44.6\% of its identities have fewer than
ten images. These properties make full-class negative interaction potentially
harmful, particularly for noisy or rarely observed classes.

\begin{table}[!t]
  \centering
  \footnotesize
  \caption{Face-verification performance across WebFace scales and class-sampling
    rates using an IR50 backbone. All result columns report ICCV21-MFR TAR (\%):
    Mask uses FAR $=10^{-4}$ and all other columns use FAR $=10^{-6}$. FC-1.0 is
    equivalent to PFC with $r=1.0$; PFC-$r$ retains every positive class and
    activates fraction $r$ of all classes. Boldface marks the highest MFR-All
    result within each WebFace scale.}
  \label{tab:webface_sampling_rates}
  \setlength{\tabcolsep}{3.8pt}
  \renewcommand{\arraystretch}{1.08}
  \resizebox{\textwidth}{!}{%
  \begin{tabular}{@{}>{\raggedright\arraybackslash}p{0.25\textwidth}*{6}{>{\centering\arraybackslash}p{0.125\textwidth}}@{}}
    \toprule
    \textbf{Training configuration} & \textbf{All} & \textbf{African} & \makecell{\textbf{Cauca-}\\\textbf{sian}} & \makecell{\textbf{South}\\\textbf{Asian}} & \makecell{\textbf{East}\\\textbf{Asian}} & \textbf{Mask}\\
    \midrule
    WF4M + FC-1.0 &86.3&83.4&91.1&88.1&65.8&72.1\\
    WF4M + PFC-0.04&74.1&71.4&81.8&76.3&52.2&54.2\\
    WF4M + PFC-0.1&85.8&83.8&91.0&87.9&66.0&71.1\\
    WF4M + PFC-0.2&86.4&84.5&91.4&88.5&66.6&71.9\\
    \textbf{WF4M + PFC-0.3}&\textbf{86.9}&84.9&91.6&88.6&67.5&72.3\\
    WF4M + PFC-0.4&86.8&84.8&91.4&88.4&67.2&72.0\\
    \midrule
    WF12M + FC-1.0&91.7&90.7&94.9&93.4&75.1&80.5\\
    WF12M + PFC-0.013&87.9&87.1&92.3&90.7&68.3&73.0\\
    WF12M + PFC-0.1&91.2&90.8&94.7&93.2&75.0&79.7\\
    WF12M + PFC-0.2&91.8&91.1&95.0&93.5&75.9&79.9\\
    \textbf{WF12M + PFC-0.3}&\textbf{91.8}&91.1&95.0&93.6&75.6&80.1\\
    WF12M + PFC-0.4&91.8&91.0&95.0&93.4&75.6&80.6\\
    \midrule
    WF42M + FC-1.0&93.9&93.3&96.2&95.2&79.5&83.9\\
    WF42M + PFC-0.008&91.3&90.3&95.2&93.0&76.9&81.2\\
    WF42M + PFC-0.1&94.0&93.5&96.4&95.5&80.0&83.8\\
    \textbf{WF42M + PFC-0.2}&\textbf{94.0}&93.7&96.4&95.5&80.1&84.3\\
    WF42M + PFC-0.3&94.0&93.7&96.4&95.5&79.8&84.5\\
    WF42M + PFC-0.4&94.0&93.4&96.4&95.5&79.6&84.4\\
    \bottomrule
  \end{tabular}
  }
\end{table}

\newcommand{\robustnesstable}{%
\begin{table}[!t]
  \centering
  \footnotesize
  \caption{Robustness under synthetic inter-class conflict, label flips, and
    long-tailed data, using an IR50 backbone. All columns report ICCV21-MFR TAR
    (\%) at FAR $=10^{-6}$. FC-1.0 is full softmax; a star denotes the conflict
    filter of Eq.~\ref{eq:noise_filtering}, which discards active negative
    logits with cosine similarity above $\tau=0.4$. The $\Delta$ column gives
    the gain over the FC-1.0 baseline of the same stress test.}
  \label{tab:webface_robustness}
  \setlength{\tabcolsep}{3.8pt}
  \renewcommand{\arraystretch}{1.08}
  \resizebox{\textwidth}{!}{%
  \begin{tabular}{@{}>{\raggedright\arraybackslash}p{0.115\textwidth}>{\raggedright\arraybackslash}p{0.115\textwidth}>{\centering\arraybackslash}p{0.085\textwidth}>{\raggedleft\arraybackslash}p{0.085\textwidth}*{4}{>{\centering\arraybackslash}p{0.125\textwidth}}@{}}
    \toprule
    \textbf{Stress test} & \textbf{Method} & \textbf{All} & \textbf{$\Delta$} & \textbf{African} & \makecell{\textbf{Cauca-}\\\textbf{sian}} & \makecell{\textbf{South}\\\textbf{Asian}} & \makecell{\textbf{East}\\\textbf{Asian}}\\
    \midrule
    Conflict & FC-1.0    & 79.9 & & 79.1 & 87.6 & 84.5 & 55.8\\
    Conflict & FC*-1.0   & 91.2 & \footnotesize{+11.3} & 90.3 & 94.5 & 92.7 & 74.4\\
    Conflict & PFC-0.1   & 91.2 & \footnotesize{+11.3} & 90.7 & 94.7 & 93.4 & 75.0\\
    Conflict & PFC*-0.1  & 91.6 & \footnotesize{+11.7} & 91.0 & 94.8 & 93.4 & 75.4\\
    Conflict & PFC*-0.2  & 91.7 & \footnotesize{+11.8} & 91.2 & 95.0 & 93.6 & 75.5\\
    Conflict & PFC*-0.3  & 91.7 & \footnotesize{+11.8} & 91.0 & 94.9 & 93.6 & 75.5\\
    \midrule
    Flip 10\% & FC-1.0   & 88.8 & & 87.1 & 92.8 & 90.6 & 70.7\\
    Flip 10\% & PFC-0.1  & 89.6 & \footnotesize{+0.8} & 89.6 & 94.0 & 92.2 & 72.2\\
    Flip 10\% & PFC*-0.1 & 90.0 & \footnotesize{+1.2} & 89.8 & 94.1 & 92.3 & 73.3\\
    \addlinespace[2pt]
    Flip 20\% & FC-1.0   & 85.4 & & 84.0 & 90.9 & 88.0 & 65.5\\
    Flip 20\% & PFC-0.1  & 87.6 & \footnotesize{+2.2} & 87.5 & 92.8 & 90.9 & 69.3\\
    Flip 20\% & PFC*-0.1 & 88.2 & \footnotesize{+2.8} & 88.0 & 93.2 & 91.2 & 70.1\\
    \addlinespace[2pt]
    Flip 40\% & FC-1.0   & 43.9 & & 41.6 & 52.8 & 48.0 & 28.6\\
    Flip 40\% & PFC-0.1  & 78.5 & \footnotesize{+34.6} & 79.3 & 87.5 & 83.9 & 57.5\\
    Flip 40\% & PFC*-0.1 & 80.2 & \footnotesize{+36.3} & 80.6 & 88.7 & 85.0 & 59.9\\
    \midrule
    Long tail & FC-1.0   & 87.4 & & 85.8 & 91.9 & 89.3 & 69.4\\
    Long tail & PFC-0.1  & 91.9 & \footnotesize{+4.5} & 90.7 & 94.8 & 92.8 & 76.2\\
    Long tail & PFC-0.2  & 92.0 & \footnotesize{+4.6} & 91.1 & 95.1 & 93.5 & 76.5\\
    Long tail & PFC-0.3  & 91.6 & \footnotesize{+4.2} & 90.6 & 94.8 & 93.3 & 76.0\\
    Long tail & PFC-0.4  & 91.0 & \footnotesize{+3.6} & 90.1 & 94.6 & 93.1 & 76.0\\
    \bottomrule
  \end{tabular}
  }
\end{table}
}
\noindent\textbf{Sampling-rate ablation.} Table~\ref{tab:webface_sampling_rates}
varies the sampling rate at three training scales while holding the backbone,
margin loss, and schedule fixed, so the only change is the fraction of activated
classes. Three regimes emerge. For $r\in[0.2,0.4]$, PFC matches or exceeds the
full-class baseline at every scale: MFR-All moves from $86.3$ to $86.9$ on
WebFace4M, from $91.7$ to $91.8$ on WebFace12M, and from $93.9$ to $94.0$ on
WebFace42M, with the same ordering on all four demographic groups; on the masked
track the ordering is less regular, with four of the nine configurations in this
range falling below their baseline. At $r=0.1$ the results are close to but
slightly below the baseline on the two smaller sets ($-0.5$ and $-0.5$ points)
and already at the baseline on WebFace42M ($94.0$ versus $93.9$). Only at the
smallest rates does accuracy degrade clearly, by $12.2$ points at $r=0.04$ on
WebFace4M, $3.8$ points at $r=0.013$ on WebFace12M, and $2.6$ points at
$r=0.008$ on WebFace42M.

Two observations follow directly from these numbers. First, the useful operating
range is broad: activating $20$--$40\%$ of the classes is sufficient at all
three scales, and no configuration in that range loses MFR-All accuracy relative
to full softmax. Second, the tolerable rate decreases as the class count grows. The
three lowest settings activate roughly $8$K, $8$K, and $16$K classes per
iteration, so the degradation is better explained by the absolute number of
active negatives than by the rate itself; $r=0.1$ corresponds to $20$K classes
on WebFace4M but $200$K on WebFace42M, which is consistent with the residual gap
appearing only at the smaller scale. We do not claim a mechanism for the mild
improvement over full softmax at $r\in[0.2,0.4]$ from this experiment alone; the
noise and long-tail ablations below isolate one contributing factor, namely
reduced exposure to conflicting and rarely observed negatives.

\noindent\textbf{Noise and data-quality ablation.} We evaluate the robustness of
PFC under three synthetic stress tests targeting inter-class conflicts, label
noise, and long-tailed identity distributions. WebFace12M-Conflict is
constructed by randomly splitting 200K identities into 600K pseudo-classes,
yielding 1M classes with extensive inter-class conflicts. For label-noise
experiments, 10\%, 20\%, or 40\% of WebFace12M image labels are randomly
replaced. For the long-tail setting, 200K WebFace42M identities are retained
unchanged, while each of the remaining 1.8M identities is randomly reduced to
two to four images. Table~\ref{tab:webface_robustness} reports the complete
results.

\robustnesstable

\emph{Inter-class conflict.} Full softmax is the weakest configuration here
($79.9$ MFR-All), because every duplicated pseudo-class acts as a negative for
its own images in every iteration. Sampling alone removes most of the damage:
PFC-0.1 reaches $91.2$, a gain of $11.3$ points, without any explicit filtering.
Adding the conflict filter at the same rate contributes a further $0.4$ points
(PFC*-0.1, $91.6$), and the best filtered configuration reaches $91.7$ at
$r=0.2$. Notably, applying the filter to full softmax (FC*-1.0, $91.2$) recovers
a comparable amount, which indicates that sampling and filtering suppress the
same failure mode rather than two independent ones.

\emph{Label noise.} The three flip levels separate the two effects. At 10\%
noise the baseline is already close to the clean result and PFC-0.1 gains only
$0.8$ points; at 20\% the gain is $2.2$ points; at 40\% the baseline collapses
to $43.9$ while PFC-0.1 reaches $78.5$, a gain of $34.6$ points. The conflict
filter adds $0.4$, $0.6$, and $1.7$ points at the three levels respectively.
Both mechanisms therefore scale with corruption severity, and the sampling term
dominates the filtering term throughout.

\emph{Long tail.} PFC-0.2 improves MFR-All from $87.4$ to $92.0$, a gain of
$4.6$ points. Here the trend reverses beyond $r=0.2$:
accuracy decreases monotonically to $91.0$ at $r=0.4$, so on long-tailed data an
intermediate rate is preferable to a larger one. This is the setting in which
increasing $r$ within $[0.1,0.4]$ is most clearly harmful, which is consistent with the
interpretation that the benefit comes from reduced exposure to rare and
unreliable negatives rather than from sampling per se.

Across all three stress tests the gains hold on every demographic group, but the
group that benefits most depends on the corruption type. Under inter-class
conflict and long-tailed data the largest gains occur on East Asian ($+19.7$ and
$+7.1$ points), the group with the lowest baseline accuracy, whereas under 40\%
label flips East Asian gains the least ($+31.3$) and African the most ($+39.0$).
Partial activation therefore does not uniformly favor the weakest group.

Across these ablations, moderate sampling rates consistently retain the accuracy
of full softmax on clean data while reducing exposure to noisy negatives. The
advantage becomes more pronounced under inter-class conflicts, label corruption,
and long-tailed identities, where PFC and PFC* substantially recover the
degradation suffered by full FC. These results indicate that partial class
activation is not only a computational approximation, but also an effective
regularizer for imperfect web-scale training data.

\section{Conclusion}
We presented Partial FC as a positive-preserving approximation to large-class
softmax: every target class is retained, while only a sampled subset of negative
classes is activated. Combined with permanent class ownership, this design
reduces classifier computation and logit memory, enabling efficient training at
massive class scales. Experiments show that moderate sampling rates maintain
competitive face-recognition accuracy and improve robustness under noisy and
long-tailed training data.

\clearpage
\bibliography{main}

\end{document}